\pdfoutput=1

\documentclass[11pt]{article}

\usepackage[final]{acl}

\usepackage{times}
\usepackage{latexsym}

\usepackage[T1]{fontenc}

\usepackage[utf8]{inputenc}

\usepackage{microtype}

\usepackage{inconsolata}

\usepackage{graphicx}

\usepackage{graphicx}      
\usepackage{subcaption}    
\usepackage{amsmath}
\usepackage{booktabs}
\usepackage{multirow}
\usepackage{array}
\usepackage{makecell}
\usepackage{hyperref}
\usepackage{amssymb}
\usepackage{afterpage}

\title{
Beyond the Next Token: 
Towards Prompt-Robust Zero-Shot Classification \\
via Efficient Multi-Token Prediction
}

\author{
    \textbf{Junlang Qian\textsuperscript{1}} \enspace
    \textbf{Zixiao Zhu\textsuperscript{1,2}} \enspace
    \textbf{Hanzhang Zhou\textsuperscript{1,2}} \enspace
    \textbf{Zijian Feng\textsuperscript{1,2}} \enspace
    \textbf{Zepeng Zhai\textsuperscript{3}} \\
    \textbf{Kezhi Mao\textsuperscript{1,2}}\thanks{Corresponding author}
\\
    \textsuperscript{1}Nanyang Technological University, Singapore\\
    \textsuperscript{2}Singapore-ETH Centre \enspace
    \textsuperscript{3}Tencent, China\\
    \texttt{
    \{junlang001,zixiao001,hanzhang001,feng0119\}@e.ntu.edu.sg}
    \\\texttt{zealzhai@tencent.com} \quad\quad\quad\quad\quad \texttt{ekzmao@ntu.edu.sg}\quad\quad\quad
}

\begin{document}
\maketitle
\begin{abstract}

Zero-shot text classification typically relies on prompt engineering, but the inherent prompt brittleness of large language models undermines its reliability. 
Minor changes in prompt can cause significant discrepancies in model performance. 
We attribute this prompt brittleness largely to the narrow focus on next-token probabilities in existing methods.
To address this, we propose \textbf{P}laceholding \textbf{P}arallel \textbf{P}rediction ($\boldsymbol{\mathcal{P}^\mathit{3}}$), a novel approach that predicts token probabilities across multiple positions and simulates comprehensive sampling of generation paths in a single run of a language model.
Experiments show improved accuracy and up to 98\% reduction in the standard deviation across prompts, boosting robustness. 
Even without a prompt, $\mathcal{P}^\mathit{3}$ maintains comparable performance, reducing the need for prompt engineering.  

\end{abstract}

\section{Introduction}

\begin{figure}[]

\centering
\setlength{\abovecaptionskip}{0.2cm} 
\includegraphics[width=1\linewidth]{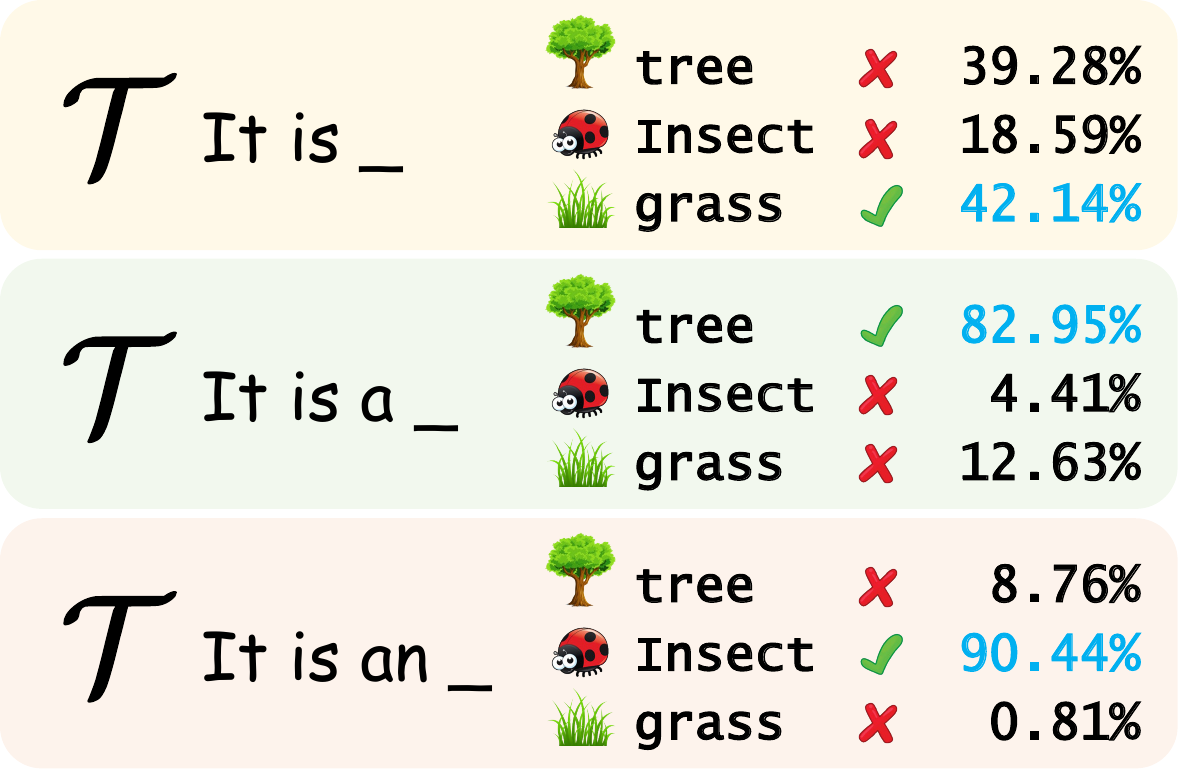}
\caption{
An example of prompt brittleness.
The prompt ``It is a \_'' yields a notably high score for ``tree'', while ``It is an \_'' overwhelmingly favors ``insect''.
The percentage scores are normalized for an arbitrary text unrelated to any class $\mathcal{T}$ = ``knows grammar.''
}
\vspace{-2mm}

\setlength{\belowdisplayskip}{0pt}
\label{fig:treegrassinsect}
\end{figure}

Zero-shot text classification \cite{radford2019language,hu2021knowledgeable,NEURIPS2022_9d560961,wang2023selfconsistency,liu2023pre,yang2023autogpt} is among the most challenging applications of pre-trained large language models (LLMs) \cite{brown2020language,touvron2023llama,anthropic2024claude}, as it aims to categorize text without additional data. 
In this context, prompt engineering has become a widely adopted approach to enhance accuracy.
However, language models exhibit inherent \textbf{prompt brittleness} \cite{sclar2023quantifying,zamfirescu2023johnny,zhou2024batchcalibrationrethinkingcalibration}: their outputs are highly sensitive to minor modifications in prompt wording or format (as shown in Figure~\ref{fig:treegrassinsect}), leading to inconsistent performance.
This issue makes crafting effective and reliable prompts difficult, particularly in zero-shot scenarios where prior information is lacking.

While some efforts have attempted to mitigate this issue through training \cite{tam2021improving}, general methods for the inference stage are rare.
Calibration \cite{zhao2021calibrate} is one of the few, but its primary benefit lies in enhancing accuracy rather than robustness, leaving the severe prompt brittleness problem not sufficiently resolved.

In this paper, we present a new perspective.  
We observe that current methods predominantly rely on next-token prediction \cite{NIPS2000_728f206c,NIPS2014_a14ac55a} to classify. 
We posit that this sole \textbf{reliance on the next token may contribute to prompt brittleness}, based on the following intuitions:
(1) Since the next token directly follows the input, it would be inevitably impacted by the prompt, whereas later tokens are likely less sensitive\footnote{
The Markov assumption implies that nearby words (or the preceding n-gram) have a stronger influence on predictions \citep{6773024,shannon1951prediction,brown1992class,almutiri2022markov}.
};
(2) Language models may not immediately provide the answer as the next token. Instead, they often first generate non-discriminative words such as ``so'', ``a'', or ``very'', or engage in preliminary reasoning\footnote{
Some examples are in appendix~\ref{sec:OutputExamples}.
}, which could hurt next-token performance for certain prompts.

\begin{figure*}[t!]
    \centering

\setlength{\abovecaptionskip}{0.2cm}

    \includegraphics[width=0.8\textwidth]{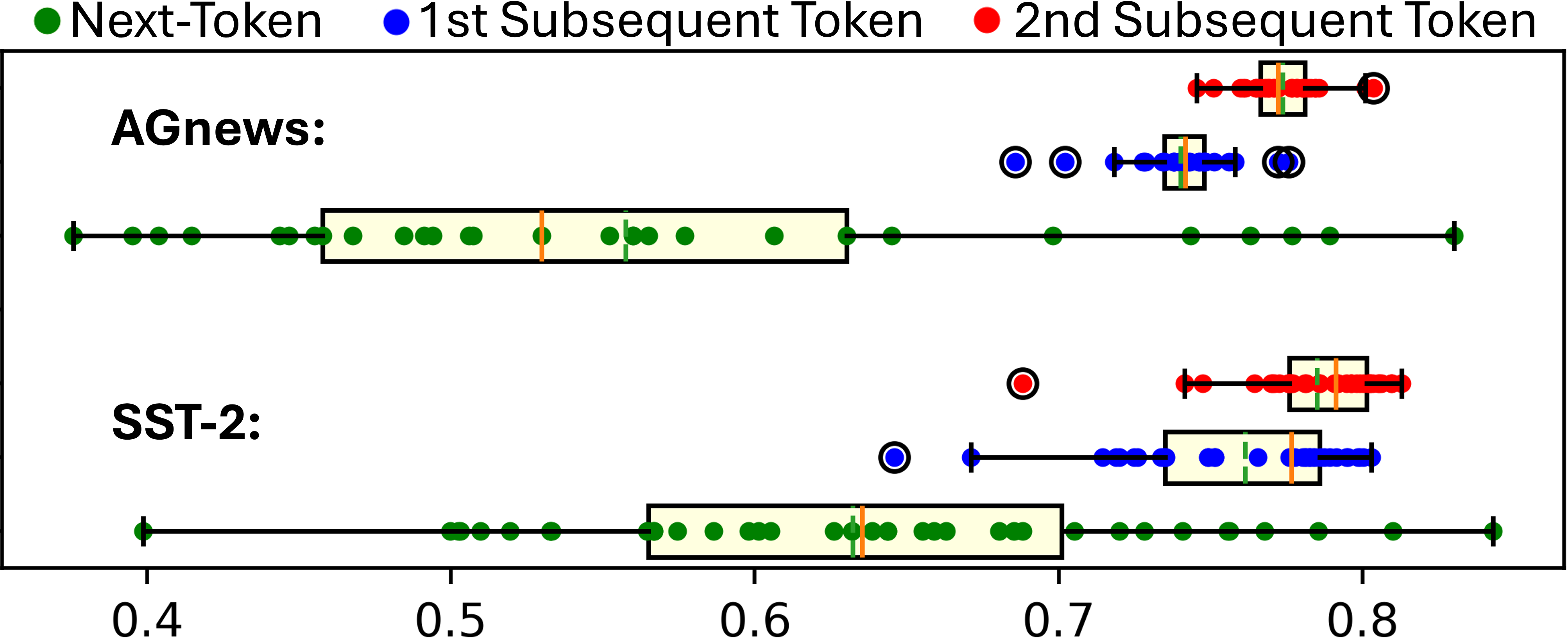}

    \caption{
Accuracies of plausible prompts using tokens at the first three positions.
Each row corresponds to a position: the \textcolor[rgb]{0.0, 0.5, 0.0}{$\bullet$} green dots represent the next token (the 0th position), and the \textcolor{blue}{$\bullet$} blue and \textcolor{red}{$\bullet$} red dots represent the 1st and 2nd tokens.
Each dot denotes a prompt, and its horizontal coordinate indicates its performance.  
    }

    \label{fig:boxplot}

    
\end{figure*}
Our experiments support our hypothesis.
As illustrated in Figure~\ref{fig:boxplot}, the accuracies across different prompts based on next-token prediction (\textcolor[rgb]{0.0, 0.5, 0.0}{$\bullet$}, green dots) are highly dispersed. 
In contrast, the accuracies for later tokens (\textcolor{blue}{$\bullet$}\textcolor{red}{$\bullet$}, blue and red dots) are more tightly clustered together and demonstrate better overall performance. 
This suggests that subsequent token predictions are more robust to prompt variations and have the potential to improve performance.

However, to unlock the potential of subsequent token predictions, we must address a key challenge: current auto-regressive \textbf{language models predict only the next token}\footnote{
An exception exists: one language model introduced by \citeauthor{gloeckle2024better} can predict up to four tokens, but is still insufficient for our needs (more than ten tokens). Therefore, this model is not included in our study.
}.

A naive solution is a token-by-token generation strategy, but it explores only a single generation path. 
Consequently, this strategy introduces error propagation, controllability issues, dependency on decoding algorithms, and randomness in generation path selection, all of which can hurt robustness\footnote{
In practice, generation-based strategies typically underperform next-token prediction in classification tasks \cite{puri2019zero,wei2021finetuned}.
}.
In zero-shot natural language generation (NLG) tasks, a popular method to enhance this strategy is to sample multiple alternative generation paths and ensemble the results \cite{wang2023selfconsistency,lin2024just,zhang2024nash}.
Although extensive sampling could theoretically outperform next-token predictions in zero-shot classification\footnote{
If all possible generation paths were enumerated, token probabilities for all positions could be calculated, encompassing and exceeding the capabilities of next-token prediction.
}, the computation of repeatedly generating numerous tokens many times is overly costly.
Thus, the \textbf{inefficiency} stemming from token-by-token generation and path dependency limits our ability to leverage subsequent-token predictions to overcome prompt brittleness.

\begin{figure}[h!]
    \centering

\setlength{\abovecaptionskip}{0.2cm} 
    
    \includegraphics[width=1\linewidth]{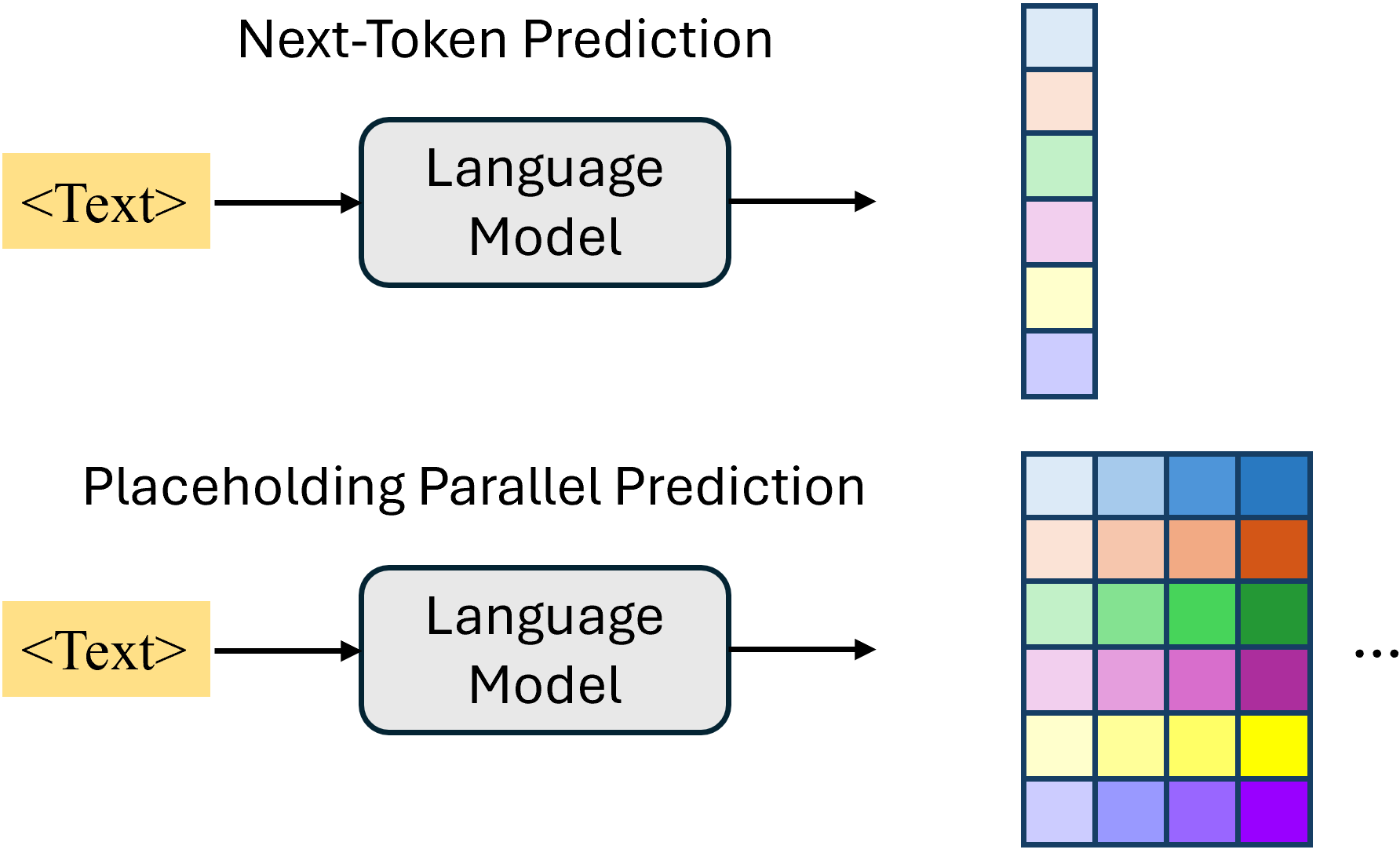}
    \caption{
Next-Token Prediction versus Placeholding Parallel Prediction.
Our proposed $ \mathcal{P}^\mathit{3} $ obtains multiple token predictions in a single language model run.
    }
    \label{fig:diagram}
\end{figure}

To address this, we propose \textbf{Placeholding Parallel Prediction} ($\boldsymbol{ \mathcal{P}^\mathit{3} }$), a novel and pluggable method.  
$\mathcal{P}^\mathit{3} $  appends placeholder tokens at the end of the input sequence to simulate comprehensive sampling of generation paths, thereby enabling multiple token predictions simultaneously within one model run.
$\mathcal{P}^\mathit{3} $ introduces subsequent token predictions into zero-shot classification with high efficiency and without being affected by generation path dependency.
This offers an effective solution to prompt brittleness.
Through extensive experiments on seven public benchmarks, we demonstrate that $\mathcal{P}^\mathit{3} $ significantly alleviates prompt brittleness and surpasses SoTA accuracy.
Notably, with $\mathcal{P}^\mathit{3} $, performance without any prompt instructions matches that with crafted prompts, significantly reducing the need for prompt engineering.

Our contributions are as follows:
\begin{itemize}
    \item 
    We propose a new perspective: the narrow focus on next-token prediction may be a key contributor to prompt brittleness in zero-shot classification.
    
    \item 
    We introduce $\mathcal{P}^\mathit{3} $, the first method to obtain multiple subsequent token predictions for text classification within a single language model run. 
    Our code is 
\href{https://github.com/qianjunlang/PlaceholdingParallelPrediction}
{open-sourced}.
    
    \item 
    Extensive experiments demonstrate that $\mathcal{P}^\mathit{3} $ significantly reduces prompt brittleness and outperforms SoTA accuracy. Notably, we find that even without any prompt, the performance is comparable to that with crafted prompts, greatly reducing the reliance on prompt engineering.

\end{itemize}

\section{Related Work}
\label{sec:related}

\subsection{Prompt Brittleness}

Prompt brittleness has emerged as a significant challenge in zero-shot classification \cite{krause2020gedi, schick2021fewshot}. This brittleness refers to the sensitivity of model performance to the specific wording and structure of prompts, where minor changes can lead to significant variations in output quality \cite{zhou2024unibias}. Training-based solutions to address prompt brittleness often rely on labeled or unlabeled data. For instance, \citet{logan2022cutting} introduced strategies for simplifying prompt engineering through finetuning, but their methods require datasets for parameter optimization. Similarly, \citet{genco2023,wang2023pesco} leveraged contrastive self-training, which involves iterative pseudo-labeling using generated data, but this also depends on large amounts of unlabeled data to work. Such methods are therefore impractical for zero-shot learning.

Consequently, few attempts have focused on mitigating prompt brittleness directly during the inference phase, where the model operates without any further adjustments. Calibration methods \cite{zhao2021calibrate} represent one of the few efforts in this area. \citeauthor{zhao2021calibrate} propose calibration strategies to normalize model outputs based on contextual priors, improving consistency across different prompts without requiring external data.
Another line of research aims to bypass prompt brittleness by automatically generating task-specific prompts \cite{pryzant2023automatic,gao2021making,holtzman2021surface, jiang2020can}. 
Rather than fundamentally addressing the brittleness problem, these approaches serve as workarounds, sacrificing interpretability and cross-task consistency.

\subsection{Zero shot Text Classification}
Since class labels typically possess clear semantics, we can leverage the capabilities of language models for zero-shot classification \cite{radford2021learning,zhou2023tent}.

Zero-shot text classification \cite{puri2019zero,brown2020language,liu2023pre} has evolved from earlier approaches like embedding-based semantic matching and natural language inference (NLI). Embedding-based methods \cite{cer2018universal,reimers2019sentencebert} suffer from semantic ambiguity, as inputs with subtle differences can receive similar embeddings. NLI-based methods \cite{bowman2015large,ma-etal-2021-issues,zhu2024edentail} decompose a multi-class problem into binary tasks, but this often results in poor calibration across binary classifiers, with inconsistent performance between classes.

Recent methods focus on next-token prediction and generation-based techniques. In next-token prediction, the model assigns a score based on the generation probability of a class label token \cite{zhao2021calibrate}. Although efficient, this method considers only a single token, limiting its capacity \cite{fengunveiling}. Generation-based methods, on the other hand, prompt the model to generate text and analyze label presence \cite{radford2019language,brown2020language,wang2023selfconsistency}. While these approaches can capture richer semantics, they are constrained to one generation path at a time and face controllability issues \cite{krause2020gedi,NEURIPS2022_b1efde53}.

Although these recent approaches address some shortcomings of earlier methods, they remain incomplete, as one focuses narrowly on a single token and the other only explores one possible output path. This leaves room for improvements in achieving more comprehensive classification strategies.

\section{Methodology}

\subsection{Background}

Text classification seeks to assign the correct label \(c^\star\) to a given text \(\boldsymbol{t}\), with \(\mathcal{C} = \{c_1, c_2, \ldots, c_k\}\) representing the predefined set of candidate categories.
Current approaches to zero-shot text classification rely on the assumption that the correct class label \(c^\star\) or its corresponding tokens will exhibit higher generation probabilities under the language model than incorrect ones:
\[
P(c^\star \mid \boldsymbol{t}) > P(c' \mid \boldsymbol{t}), \quad \forall c' \in \mathcal{C}, \, c' \neq c^\star.
\]
A language model $\mathcal{LM}$ predicts the generation probability of the next token given an input sequence \( \boldsymbol{x} \). Specifically,  
\[
    \mathcal{LM}(\boldsymbol{x})_{x_{n}} = P(x_{n} \mid \left(x_0, x_1, \ldots, x_{n-1}\right)),
\]
where the input \( \boldsymbol{x} \) is constructed by integrating the given text \( \boldsymbol{t} \) into a prompt template \( p(\cdot) \): 
\[
\boldsymbol{x} = (x_0, x_1, \ldots, x_{n-1}) = p(\boldsymbol{t}).
\]

The most commonly used approach is \textbf{based on next token prediction}, which assumes that the class probability \( P(c \mid \boldsymbol{t}) \) is approximated by the language model's next token probability:
\[
P(c \mid \boldsymbol{t}) \approx \mathcal{LM}(\boldsymbol{x})_{x_{n} = c}.
\]
Therefore, the class label is predicted by:
\[
\hat{c} \leftarrow \arg\max_{c \in \mathcal{C}} \mathcal{LM}(\boldsymbol{x})_{x_{n} = c}\text{.}
\]

An alternative approach \textbf{based on generation} leverages the continuation capability of language models to generate an output sequence:
\[
\boldsymbol{x} \xrightarrow{\mathcal{LM}} \hat{(x_{n}, x_{n+1}, \ldots, x_{n+m-1})}.
\]
The class label is then directly determined by which one appears in the generated sequence:
\[
\hat{c} \leftarrow \left\{\hat{x}_{n}, \hat{x}_{n+1}, \ldots, \hat{x}_{n+m-1}\right\} \cap \mathcal{C}.
\]

However, both methods remain incomplete: next-token prediction focuses on a single token position, while generation-based methods consider only one instance among many possible generation paths, limiting their robustness.

\subsection{Placeholding Skipping Prediction ($\mathcal{PSP}$)}

Let the $i$-th output token be $x_{n+i}$ (where the next token is the 0-th token).  
All possible generation prefixes can be expressed as:  
$$
\left\{\left(x_n, x_{n+1}, \ldots, x_{n+i-1}\right) \mid x_j \in \mathcal{V}, n \leq j<n+i\right\},
$$
where \(\mathcal{V}\) denotes the vocabulary, and \(\boldsymbol{x}\) represents the known input.

The probability of each prefix is:  
\[
P\left(x_n, x_{n+1}, \ldots, x_{n+i-1} \mid \boldsymbol{x}\right).
\]

Given this prefix, the probability of \(x_{n+i} = c\) is:  
\[
P\left(x_{n+i} = c \mid \boldsymbol{x}, x_{n+1}, \ldots, x_{n+i-1} \right).
\]

Thus, the probability of the entire generation path is:  

$P\left(x_n, x_{n+1}, \ldots, x_{n+i}=c \mid \boldsymbol{x}\right)$
\[
\begin{aligned}
&= P\left(x_n, x_{n+1}, \ldots, x_{n+i-1} \mid \boldsymbol{x}\right) \\
&\quad \times P\left( x_{n+i} = c \mid \boldsymbol{x}, x_{n+1}, \ldots, x_{n+i-1} \right).
\end{aligned}
\]

Next, by enumerating all possible prefixes, the overall probability for \(x_{n+i} = c\) becomes: 
\begin{multline}
      P\left(x_{n+i} = c \mid \boldsymbol{x}\right) 
 \\
 = \sum_{\substack{(x_{n}, \ldots, x_{n+i-1}) \\ \in \mathcal{V}^{i}}} 
P\left(x_n, x_{n+1}, \ldots, x_{n+i-1} \mid \boldsymbol{x}\right) \\ \nonumber
\times P\left(x_{n+i} = c \mid \boldsymbol{x}, x_n,x_{n+1}, \ldots, x_{n+i-1}\right). \nonumber
\end{multline}

This is equivalent to:  
\begin{multline}
    P\left(x_{n+i} = c \mid \boldsymbol{x}\right) \\ 
 = P\left(x_{n+i} = c \,\middle|\, \boldsymbol{x}, 
\underbrace{x_n, x_{n+1}, \ldots, x_{n+i-1}}_{\text{unknown}} \right). \nonumber
\end{multline}

We approximate this as: 
\begin{multline}
    P\left(x_{n+i} = c \mid \boldsymbol{x}\right) \\
\approx P\left(x_{n+i} = c \,\middle|\, \boldsymbol{x}, 
\underbrace{\texttt{<ph>}, \texttt{<ph>}, \ldots, \texttt{<ph>}}_{\text{i times}} \right). \nonumber
\end{multline}

This can be rewritten using the language model as:
\begin{multline}
    P\left(x_{n+i} = c \mid \boldsymbol{x}\right) \\
\approx \mathcal{LM}\left(\boldsymbol{x}, 
\underbrace{\texttt{<ph>}, \texttt{<ph>}, \ldots, \texttt{<ph>}}_{\text{i times}}\right)_{x_{n+i}=c}. \nonumber
\end{multline}

\begin{figure}[b!]


\centering

\includegraphics[width=1\linewidth]{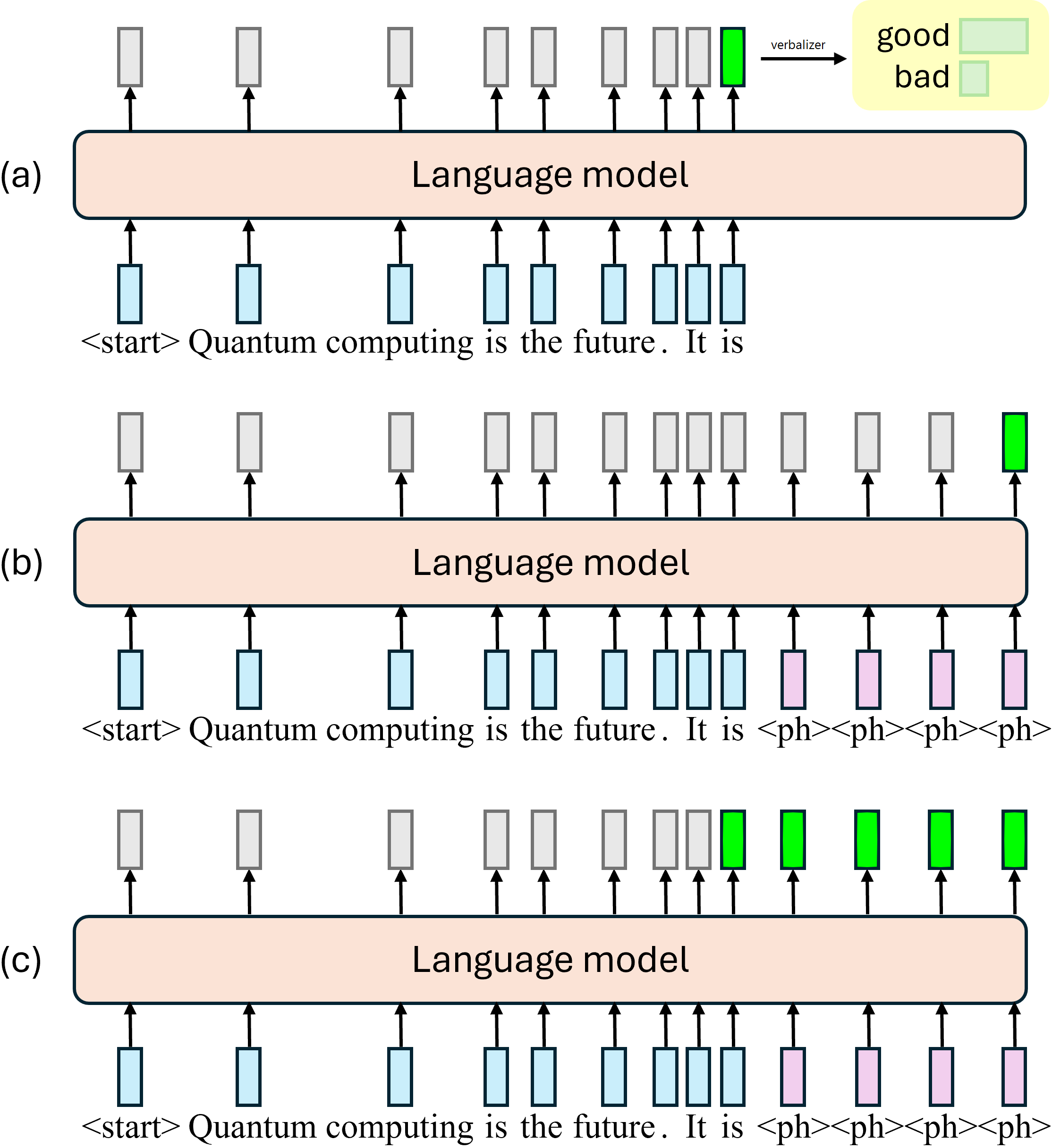}

\caption{
(a) Next-Token Prediction. (b) Placeholding Skipping Prediction ($\mathcal{PSP}$). (c) Placeholding Parallel Prediction ($\mathcal{P}^\mathit{3}$). 
The small green rectangles indicate the output tokens to be used, and the grey ones indicate those not to be used. 
\texttt{<ph>} represents a placeholder token.
}

\setlength{\belowdisplayskip}{0pt}
\label{fig:methodfig}
\end{figure}

As shown in Figure~\ref{fig:methodfig}(b), we append \( i \) placeholder tokens, \texttt{<ph>}\footnote{
In LLaMA2, we selected the unknown token \texttt{<unk>} as the placeholder \texttt{<ph>}.
This token originally signifies "unknown", representing out-of-vocabulary (OOV) tokens, such as unrecognized language or unreadable characters, which can somewhat convey the intended meaning.
Additionally, \texttt{<unk>} offers two key advantages: (1) it has length, and (2) it carries no semantic meaning.
}, to the input sequence \( \boldsymbol{x} \) and feed this extended sequence into the language model. This allows us to obtain the prediction for the \( i \)-th subsequent token, formally expressed as:
\[
\mathcal{PSP}( \boldsymbol{x}, i) = \mathcal{LM}(\boldsymbol{x'})
\text{,}
\]
where:
\[
\boldsymbol{x'} =
(
x_0, x_1, \ldots, 
x_{n-1}, \underbrace{\texttt{<ph>}, \texttt{<ph>}, \ldots, \texttt{<ph>}}_{i \text{ times}} 
)
\text{.}
\]
We refer to this approximation of the $i$-th token prediction as Placeholding Skipping Prediction ($ \mathcal{PSP}$).

\subsection{Placeholding Parallel Prediction ($ \mathcal{P}^\mathit{3} $)}
\label{sec:P3}

In practical implementation, LLMs employ a transformer architecture with blocks stacked in both depth and sequence length, maintaining width-aligned input and output.
Transformers with unidirectional attention produce an output for each prefix of the input sequence \( \boldsymbol{x} \) during inference:
\[
\text{transformer}(\boldsymbol{x}) = \left[ 
\begin{aligned}
& \mathcal{LM}(\varnothing), \\ 
& \mathcal{LM}(x_0), \\ 
& \mathcal{LM}(x_0, x_1), \\
& \ldots, \\ 
& \mathcal{LM}(x_0, x_1, \ldots, x_{n-1}) 
\end{aligned}
\right]\text{.}
\]
This property supports tasks such as computing text likelihood and parallel training.  
As depicted in Figure~\ref{fig:methodfig}(a), current classification methods merely use the next-token prediction, \(\mathcal{LM}(\boldsymbol{x})=\mathcal{LM}(x_0, x_1, \ldots, x_{n-1})\), the last element of the transformer output.

Appending a series of <ph> tokens (subject to memory constraints) to the input sequence enables the transformer to compute all prefixes automatically:
\[
\text{transformer}(\boldsymbol{x'}) \quad\quad\quad\quad\quad\quad\quad\quad\quad\quad\quad
\]
\begin{equation}
\notag
\resizebox{0.82\linewidth}{!}{$
    = \left[
    \begin{aligned}
    & \mathcal{LM}(\varnothing), \\ 
    & \mathcal{LM}(x_0), \\ 
    & \mathcal{LM}(x_0, x_1), \\
    & \ldots, \\ 
    & \mathcal{LM}(x_0, x_1, \ldots, x_{n-1}), \\
    & \mathcal{LM}(x_0, x_1, \ldots, x_{n-1}, \texttt{<ph>}), \\
    & \mathcal{LM}(x_0, x_1, \ldots, x_{n-1}, \texttt{<ph>}, \texttt{<ph>}), \\
    & \ldots \\
    \end{aligned}
    \right]\text{,}
$
}
\end{equation}
where:
\[
\boldsymbol{x'} = (x_0, x_1, \ldots, x_{n-1}, \texttt{<ph>}, \texttt{<ph>}, \ldots)
\text{.}
\]
As illustrated in Figure~\ref{fig:methodfig}(c), extracting our desired elements, which correspond to the next-token and all subsequent tokens, we define:
\begin{equation}
\centering
\notag
\begin{aligned}
\mathcal{P}^\mathit{3}(\boldsymbol{x}) 
&= 
\left[ 
    \begin{aligned}
    & \mathcal{LM}(x_0, x_1, \ldots, x_{n-1}), \\ 
    & \mathcal{LM}(x_0, x_1, \ldots, x_{n-1}, \texttt{<ph>}), \\ 
    & \mathcal{LM}(x_0, x_1, \ldots, x_{n-1}, \texttt{<ph>}, \texttt{<ph>}), \\
    & \ldots 
    \end{aligned}
\right]
\\
&= 
\left[ 
    \begin{aligned}
    & \mathcal{LM}(\boldsymbol{x}), \\ 
    & \mathcal{PSP}(\boldsymbol{x}, 1), \\ 
    & \mathcal{PSP}(\boldsymbol{x}, 2), \\
    & \ldots
    \end{aligned}
\right]
= 
\left[ 
    \begin{aligned}
    & \mathcal{PSP}(\boldsymbol{x}, 0), \\ 
    & \mathcal{PSP}(\boldsymbol{x}, 1), \\ 
    & \mathcal{PSP}(\boldsymbol{x}, 2), \\
    & \ldots
    \end{aligned}
\right]
\text{.}
\end{aligned}
\end{equation}

By leveraging the transformer's inherent capability to handle multiple token predictions\footnote{
Note that our multi-token prediction does not generate a specific sequence of tokens; rather, it produces a probability distribution over the vocabulary for each token position.
} in a single run, this approach significantly enhances efficiency.
We term this method  $\mathcal{P}^\mathit{3}$.

\section{Experiment}


\begin{table*}[h!]
    \centering

    \resizebox{0.97\textwidth}{!}{ 
    \begin{tabular}{
        l|cc|cc|cccc|c
    }
    \toprule
        \multicolumn{3}{c|}{\textbf{Dataset}} & {\textbf{Gen}}& {\textbf{3w-SC}}  & {\textbf{NTP}} & \textbf{Cali1} & \textbf{Cali2} & \textbf{Cali3} & {\textbf{Ours}} \\

    \midrule
    \multirow{4}{*}{IMDb} & \multirow{2}{*}{13B} & acc (\%) & \multirow{2}{*}{--} & \multirow{2}{*}{--} & 68.11 & 79.36 & 77.92 & 75.08 & \textbf{82.35 \footnotesize +14.24} \\
    & & std & & & 0.1003 & 0.0899 & 0.0907 & 0.0962 & \textbf{0.0020 \footnotesize -98.05\%} \\
    \cmidrule{2-10}
   & \multirow{2}{*}{70B} & acc (\%) & 47.92 & 53.13 & 61.60 & 71.23 & 71.82 & 68.99 & \textbf{75.86 \footnotesize +14.26} \\
    & & std & 0.2053 & 0.3291 & 0.0729 & 0.0798 & 0.0749 & 0.0834 & \textbf{0.0721 \footnotesize -7.92\%} \\
    
    \midrule
    \multirow{4}{*}{AGnews} & \multirow{2}{*}{13B} & acc (\%) & \multirow{2}{*}{--} & \multirow{2}{*}{--} & 55.75 & 60.40 & 61.70 & 61.15 & \textbf{80.51 \footnotesize +24.76} \\
    & & std & & & 0.1267 & 0.1361 & 0.1106 & 0.1053 & \textbf{0.0068 \footnotesize -94.64\%} \\
    \cmidrule{2-10}
    & \multirow{2}{*}{70B} & acc (\%) & 58.59 & 50.88 & 48.24 & 49.37 & 50.38 & 49.62 & \textbf{80.14 \footnotesize +31.90} \\
    & & std & 0.1180 & 0.3016 & 0.1544 & 0.1546 & 0.1398 & 0.1450 & \textbf{0.0329 \footnotesize -78.71\%} \\
    
    \midrule
    \multirow{4}{*}{DBpedia} & \multirow{2}{*}{13B} & acc (\%) & \multirow{2}{*}{--} & \multirow{2}{*}{--} & 64.39 & 56.52 & 61.91 & 60.04 & \textbf{64.47 \footnotesize +0.08} \\
    & & std & & & 0.0965 & 0.2223 & 0.0922 & 0.0948 & \textbf{0.0052 \footnotesize -94.64\%} \\
    \cmidrule{2-10}
    & \multirow{2}{*}{70B} & acc (\%) & 66.41 & 56.35 & 70.03 & 67.11 & 67.23 & 65.74 & \textbf{72.98 \footnotesize +2.95} \\
    & & std & 0.0438 & 0.3288 & 0.0744 & 0.0694 & 0.0822 & 0.0865 & \textbf{0.0469 \footnotesize -36.92\%} \\

    \midrule
    \multirow{4}{*}{Amazon} & \multirow{2}{*}{13B} & acc (\%) & \multirow{2}{*}{--} & \multirow{2}{*}{--} & 65.35 & 77.37 & 78.03 & 78.62 & \textbf{81.62 \footnotesize +16.27} \\
    & & std & & & 0.0986 & 0.1086 & 0.0886 & 0.0839 & \textbf{0.0065 \footnotesize -93.45\%} \\
    \cmidrule{2-10}
     & \multirow{2}{*}{70B} & acc (\%) & 49.22 & 53.32 & 61.14 & 60.50 & 61.02 & 61.41 & \textbf{78.72 \footnotesize +17.58} \\
    & & std & 0.1413 & 0.3315 & 0.0466 & 0.0766 & 0.0699 & 0.0772 & 0.0644 \footnotesize +38.39\% \\
    
    \midrule
    \multirow{4}{*}{ISEAR} & \multirow{2}{*}{13B} & acc (\%) & \multirow{2}{*}{--} & \multirow{2}{*}{--} & 36.79 & 27.48 & 41.07 & 37.73 & \textbf{41.61 \footnotesize +4.82} \\
    & & std & & & 0.0935 & 0.1771 & 0.0712 & 0.0960 & \textbf{0.0182 \footnotesize -80.51\%} \\
    \cmidrule{2-10}
      & \multirow{2}{*}{70B} & acc (\%) & 24.09 & 25.88 & 36.90 & 40.59 & 33.30 & 37.17 & \textbf{43.22 \footnotesize +6.32} \\
    & & std & 0.0208 & 0.1500 & 0.0935 & 0.0835 & 0.0716 & 0.0875 & \textbf{0.0653 \footnotesize -30.18\% } \\
    
    \midrule
    \multirow{4}{*}{SST-2} & \multirow{2}{*}{13B} & acc (\%) & \multirow{2}{*}{--} & \multirow{2}{*}{--} & 63.21 & 73.15 & 69.52 & 67.60 & \textbf{80.14 \footnotesize +16.92} \\
    & & std & & & 0.0997 & 0.1145 & 0.0995 & 0.1131 & \textbf{0.0178 \footnotesize -82.17\%} \\
    \cmidrule{2-10}
    & \multirow{2}{*}{70B} & acc (\%) & 43.49 & 47.85 & 57.30 & 66.64 & 66.98 & 60.57 & \textbf{71.07 \footnotesize +13.77} \\
    & & std & 0.2419 & 0.3224 & 0.0612 & 0.1041 & 0.1007 & 0.0994 & 0.0843 \footnotesize +37.65\% \\
    
    \midrule
    \multirow{4}{*}{Yahoo} & \multirow{2}{*}{13B} & acc (\%) & \multirow{2}{*}{--} & \multirow{2}{*}{--} & 41.90 & 34.94 & 45.12 & 45.48 & \textbf{50.50 \footnotesize +8.59} \\
    & & std & & & 0.0967 & 0.1918 & 0.0797 & 0.0781 & \textbf{0.0073 \footnotesize -92.43\%} \\
    \cmidrule{2-10}
   & \multirow{2}{*}{70B} & acc (\%) & 36.98 & 31.93 & 37.74 & 38.85 & 40.30 & 39.07 & \textbf{50.74 \footnotesize +13.00} \\
    & & std & 0.0516 & 0.1931 & 0.1210 & 0.0980 & 0.1006 & 0.0844 & \textbf{0.0544 \footnotesize -55.03\%} \\

    \midrule
    \midrule
    \multirow{4}{*}{\textbf{Avg}} & \multirow{2}{*}{13B} & acc (\%) & \multirow{2}{*}{--} & \multirow{2}{*}{--} & 56.50 & 58.46 & 62.18 & 60.81 & \textbf{68.74 \footnotesize +12.24} \\
    & & std & & & 0.1017 & 0.1486 & 0.0904 & 0.0953 & \textbf{0.0091 \footnotesize -91.05\%}\\
    \cmidrule{2-10}
     & \multirow{2}{*}{70B} & acc (\%) & 46.67 & 45.62 & 53.28 & 56.33 & 55.78 & 54.65 &  \textbf{67.53 \footnotesize +14.25}   \\
    &  & std & 0.1175 & 0.2795 & 0.0899 & 0.0951 & 0.0914 & 0.0948 & \textbf{0.0600 \footnotesize -33.26\%}\\
    
    \bottomrule
    \end{tabular}}

    \caption{
Results of accuracy and cross-prompt standard deviation (i.e., prompt brittleness) for each dataset. 
Gen refers to a vanilla generative method, and 3w-SC denotes three-way self-consistency. 
NTP represents the vanilla next-token prediction method, which serves as our baseline. 
Cali1, Cali2, and Cali3 correspond to calibration methods using "N/A," an empty string, and "<unk>*5" as the calibration text, respectively. 
Avg represents the average across datasets.
For further details on hyperparameter settings, see appendix~\ref{sec:eta}.
As shown, our method significantly improves stability while achieving the highest accuracy. 
    }


    \label{tab:maintable}
\end{table*}

\subsection{Datasets, Models, and Prompts}

We conducted extensive experiments on two large language models and seven publicly available datasets, using over 30 plausible prompts for each dataset.

We employed seven publicly available text classification datasets: 
Amazon Review Polarity \cite{zhang2015character}, SST2 \cite{socher2013recursive}, and IMDb \cite{maas-etal-2011-learning} for sentiment classification (binary labels: positive/negative) of product and movie reviews; 
AGnews \cite{NIPS2015_250cf8b5} and DBpedia \cite{BIZER2009154} for topic classification (four and fourteen categories, respectively) of news articles and Wikipedia content; 
Yahoo Answers \cite{NIPS2015_250cf8b5} comprises questions and answers classified into ten categories. 
ISEAR \cite{scherer1994evidence} consists of sentences labeled with seven emotions for emotion classification. 
To save time, we selected 10,000 random instances from the IMDb, Amazon, DBpedia, and Yahoo datasets for LLaMA2-70b, and from the Amazon dataset for LLaMA2-13b.

The two models are LLaMA2-13b, with 13 billion parameters for efficient NLP tasks, and LLaMA2-70b, with 70 billion parameters for more complex tasks and broader context handling \cite{touvron2023llama}.

We employed a method similar to that of \citet{gonen2022demystifying} to generate prompts, using ChatGPT-4 \cite{openai2024gpt4} to expand a set of seed prompts.
A comprehensive list of all prompts used in our experiments is provided in the appendix~\ref{promptlist}.

\subsection{Comparison Methods}

Our baseline method employs the straightforward next-token prediction for classification. 
Additionally, we implemented a calibration method \cite{zhao2021calibrate} to compare its performance with subsequent tokens. 
The calibration method adjusts the conditional generation probability (classification scores) of the next-token under different prompts by introducing meaningless text inputs and performing a single additional run of the model. 
This approach achieves state-of-the-art performance in zero-shot classification without requiring any manually collected external information, knowledge, or unlabeled data resources, thereby ensuring interpretability. The calibration method's enhancements to zero-shot classification focus on two critical aspects: average accuracy and stability, both of which are pivotal to our evaluation criteria.
We used three different meaningless strings for calibration to ensure a comprehensive comparison.
We also evaluated 256 samples per dataset using generative methods and three-way self-consistency to demonstrate our efficiency.

\subsection{Experimental Details}

We conducted the experiments on two A6000 GPUs.

In our experiments, for each data sample and prompt combination, we appended 512 additional \texttt{<ph>} tokens\footnote{
The use of so many <ph> tokens was initially intended to ensure redundancy but turned out to be unnecessary.
} to examine the effectiveness of our method across a broad range. 
To demonstrate that our approach is not reliant on this specific number, we introduced the hyperparameter $\eta$ to control the range of tokens considered dynamically.
Small $\eta$ values focus on tokens close to the next position, while large $\eta$ values consider tokens far down the sequence. 
Specifically, when $\eta=0$, the method is equivalent to next-token prediction.

The two models exhibit behavior differences\footnote{
We discuss the findings of behavioral differences between the two models in appendix~\ref{modeldiff}, but since they are orthogonal to this study, we do not elaborate on them here.
}, prompting us to adapt our method accordingly. 
For LLaMA2-13B, classification is based on the probability distribution of the token located at a position proportional to the number of input tokens, determined by a slope of 
$\tan(\eta)$. 
We set $\eta$ as the angle and map it into the tangent space for smoother transitions and scaling.
For LLaMA2-70b, we define $[ 0, \eta)$ as a fixed range to vote and apply calibration to enhance the results.

\section{Results Analysis}

\subsection{Overview}
Aggregating results across seven datasets reveals that our $\mathcal{P}^\mathit{3}$ method substantially improves \textbf{efficiency}, \textbf{robustness}, and \textbf{accuracy}.

\begin{table}[h!]
\centering
\resizebox{1\linewidth}{!}{
\setlength{\tabcolsep}{8pt} 
\begin{tabular}{l|cccc}
\toprule
\multirow{2}{*}{\textbf{Dataset}} & \multicolumn{4}{c}{\textbf{Number of Runs}} \\
\cmidrule{2-5}
& \textbf{Gen} & \textbf{3w-SC} & \textbf{NTP} & $\boldsymbol{\mathcal{P}^\mathit{3}}$ \\ \midrule
Amazon  & 30.75 & 92.26  & 1 & 1 \\
IMDb    & 28.03 & 84.09  & 1 & 1 \\
AGnews  & 46.98 & 140.94 & 1 & 1 \\
DBpedia & 36.09 & 108.27 & 1 & 1 \\
Yahoo   & 34.52 & 103.55 & 1 & 1 \\
SST-2   & 25.80 & 77.39  & 1 & 1 \\
ISEAR   & 26.86 & 80.58  & 1 & 1 \\ 
\bottomrule
\end{tabular}}
\caption{
The average number of output tokens used to reach an answer (number of model runs).
For outputs exceeding 50 tokens without matching any of the options, the sequence is truncated at 50.
Gen denotes the vanilla generative approach, 3w-SC denotes 3-way self-consistency, NTP denotes next-token prediction, and $\mathcal{P}^\mathit{3}$ is ours.
}
\label{tab:efficiency}
\end{table}

\begin{figure*}[t!]
\setlength{\abovecaptionskip}{0.2cm} 
    \centering
    \begin{minipage}[b]{0.45\linewidth}
        \centering
        \includegraphics[width=\linewidth]{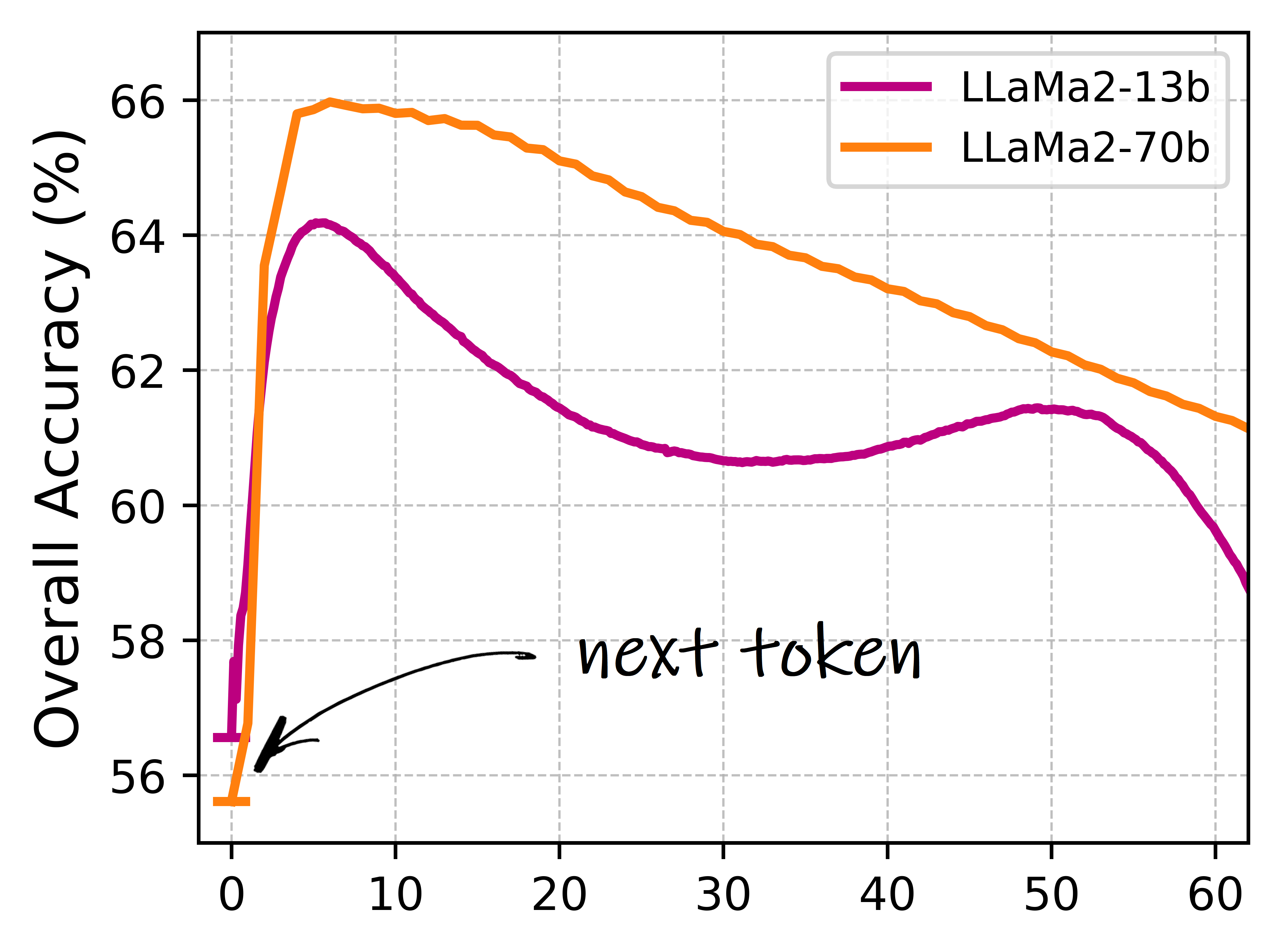}
    \end{minipage}
    \hspace{0.05\linewidth}
    \begin{minipage}[b]{0.45\linewidth}
        \centering
        \includegraphics[width=\linewidth]{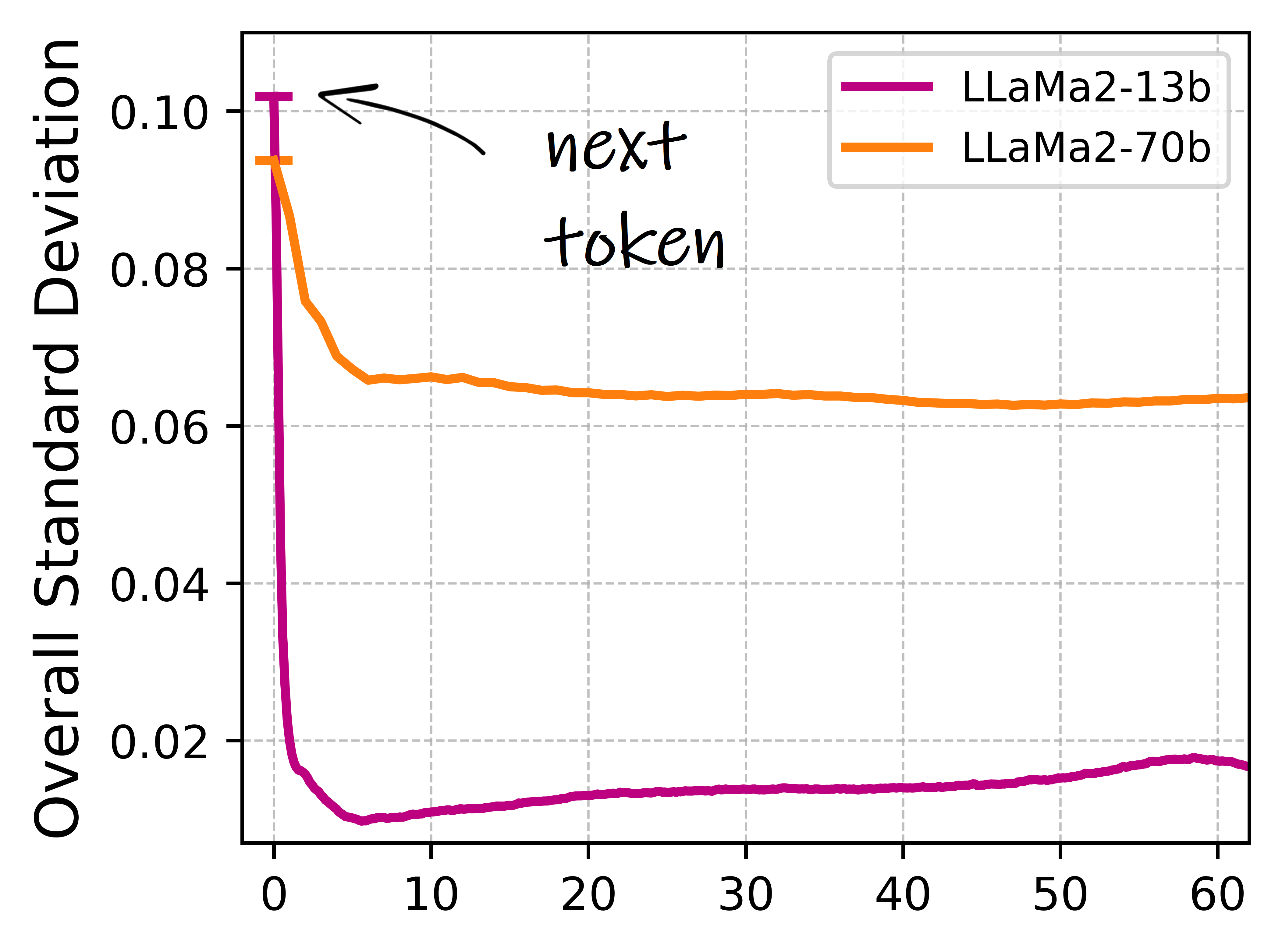}
    \end{minipage}
    \caption{
Average accuracy and average cross-prompt standard deviation (i.e., prompt brittleness) across seven datasets of $\mathcal{P}^\mathit{3}$. 
The horizontal axis represents the hyperparameter $\eta$, where $\eta=0$ corresponds to the next-token prediction results, and larger $\eta$ values indicate consideration of more distant token positions.    
$\eta>0$ shows higher accuracy and lower standard deviation compared to next-token prediction.
    }
    \label{fig:side_by_side}
\end{figure*}

\textbf{Efficiency:} Unlike generative approaches that require sequential token-by-token predictions through multiple model runs, $\mathcal{P}^\mathit{3}$ obtains multiple token predictions simultaneously in a single run. 
As evidenced in Table~\ref{tab:efficiency}, $\mathcal{P}^\mathit{3}$ achieves time complexity on par with direct next-token prediction, ensuring high efficiency.
As a trade-off for multiple token prediction, $\mathcal{P}^\mathit{3}$ consumes slightly more floating-point operations (FLOP) compared to next-token prediction, as shown in Table~\ref{tab:minortradoff}.

\begin{table}[h]
    \centering
    \resizebox{1\linewidth}{!}{
\setlength{\tabcolsep}{9pt} 
    \begin{tabular}{l|cc|cc}
    \toprule
        \multirow{2}{*}{\textbf{Dataset}} & \multicolumn{2}{c|}{\textbf{13B}} & \multicolumn{2}{c}{\textbf{70B}} \\
        \cmidrule{2-5}
        & \textbf{NTP} & $\boldsymbol{\mathcal{P}^\mathit{3}}$ & \textbf{NTP} & $\boldsymbol{\mathcal{P}^\mathit{3}}$  \\ \midrule
        Amazon & 2.82 & 3.05 & 14.26 & 14.91 \\ 
        IMDb & 8.31 & 9.03 & 41.91 & 42.56 \\ 
        AGnews & 1.62 & 1.75 & 8.20 & 8.85 \\
        DBpedia & 2.16 & 2.33 & 10.94 & 11.59 \\ 
        Yahoo & 3.67 & 3.98 & 18.56 & 19.21 \\ 
        SST-2 & 0.67 & 0.71 & 3.38 & 4.02 \\ 
        ISEAR & 0.70 & 0.75 & 3.57 & 4.22 \\ 
        \bottomrule
    \end{tabular}}
    \caption{
Average numbers of FLOPs for next-token prediction (NTP) and $\mathcal{P}^\mathit{3}$ on different datasets for 13B and 70B models, all reported in Tflops, with $\eta = 5$. 
See appendix~\ref{moreflops} for the estimation method and FLOPs for different numbers of <ph> tokens.
    }
    \label{tab:minortradoff}
\end{table}

\textbf{Robustness and Accuracy:} $\mathcal{P}^\mathit{3}$ significantly reduces prompt brittleness and achieves better performance. 

Figure~\ref{fig:side_by_side} illustrates that applying a unified hyperparameter $\eta$ to all datasets, $\mathcal{P}^\mathit{3}$ \textbf{consistently outperforms} next-token prediction ($\eta=0$) in overall accuracy and robustness within a broad range ($0<\eta<60$). 
Notably, the average cross-prompt standard deviation (i.e., prompt brittleness) reaches its minimum around $\eta=5$ and stabilizes afterward, with accuracy peaking around the same point. 
This behavior implies that $\eta\approx 5$ can serve as a nice practical default setting when without task-specific prior knowledge.

Our method adapts effectively to diverse specific scenarios, and the results for individual datasets are presented in Table~\ref{tab:maintable}. 
$\mathcal{P}^\mathit{3}$ achieves high accuracy and robustness on the evaluated datasets, outperforming baseline and state-of-the-art approaches.
Especially on LLaMA2-13b, our method achieved over 80\% reduction in standard deviation on every dataset, with an average reduction of 91\%. For AGNews, the LLaMA2-13b and LLaMA2-70b models achieved 25\% and 32\% increases in accuracy, along with 94\% and 79\% reductions in standard deviation, respectively.
On IMDb, the LLaMA2-13B model not only improved accuracy by 14\% but also yielded a standard deviation of only 0.002 (with a decrease over 98\%), almost eliminating prompt brittleness.

\subsection{Performance without Prompt}
\label{noprompt}

\begin{table}[h]
    \centering
    \resizebox{1\linewidth}{!}{
    \begin{tabular}{l|cc|cc}
    \toprule
        \multirow{2}{*}{\textbf{Dataset}} & \multicolumn{2}{c|}{\textbf{13B}} & \multicolumn{2}{c}{\textbf{70B}} \\
        \cmidrule{2-5}
        & \textbf{Crafted} & \textbf{NoP} & \textbf{Crafted} & \textbf{NoP}  \\ \midrule
        IMDb & 82.34 & 82.51  & 75.82 & 77.33  \\ 
        AGnews & 80.52 & 80.26  & 80.15 & 79.85\\ 
        Yahoo & 50.47 & 51.06 & 50.84 & 48.04 \\
        Amazon & 81.61 & 81.78  & 79.66 & 80.60  \\ 
        SST-2 & 80.13 & 80.45 & 71.23 & 65.28  \\ 
        DBpedia & 64.50 & 63.52  & 72.98 & 72.96  \\ 
        ISEAR & 41.61 & 41.66  & 43.23 & 42.66  \\

        \bottomrule
    \end{tabular}}

\caption{
The average accuracy comparison between null prompts (i.e., using no prompt) and crafted prompts. Crafted represents crafted prompts, and NoP indicates using no prompt.
}
    \label{tab:nullprompttable}
\end{table}

Given the significant reduction in standard deviation, the differences between various prompts have been largely minimized, making the choice of a specific prompt less critical. 
As shown in Table~\ref{tab:nullprompttable}, using a null prompt (i.e., without any prompt) for zero-shot classification yields performance comparable to that of crafted prompts. This demonstrates that $\mathcal{P}^\mathit{3}$ effectively mitigates prompt brittleness and substantially reduces the reliance on prompt engineering.

\section{Disscusion}

\noindent
\textbf{What is the motivation of prompt engineering?}
The need for prompt engineering arises from the prompt brittleness (or cross-prompt instability), as not all prompts yield the same accuracy.
Prompt engineering would become unnecessary if all prompts achieve consistent accuracy.
Addressing the issue of prompt brittleness would save significant costs associated with prompt engineering, making it a worthwhile research topic.

\noindent
\textbf{Is prompt engineering necessary in zero-shot classification?}
We leverage the explicit semantic information contained within the classification labels to perform zero-shot classification. We assign a sample to the label with which it has the highest semantic alignment.
We use the token generation probability as classification scores in zero-shot classification.
This approach assumes that if the context is closely related to the semantics of a certain class label, tokens associated with that class will exhibit a higher co-occurrence probability compared to tokens related to other classes.
The purpose of adding a prompt is to guide the model to generate class-related tokens as the next token. These class-related tokens may not necessarily appear at the next position but could emerge later in the text. 
The misalignment between the classification task and NLG, leading to the out-of-distribution (OOD) problem, undermines zero-shot performance by causing LLMs to generate non-discriminative words before the relevant class labels.
Some studies, like \citet{gonen2022demystifying} and \citet{zhou2023large}, alleviate the OOD issue by calculating prompt perplexity, but we address the core principle directly.
Once subsequent tokens are predicted, the prompt's influence reduces.
Logically, the text and the categories should suffice for classification. 
However, next-token predictions are often unstable, with accuracy fluctuations exceeding 10\%, leading to the introduction of prompt engineering. Subsequent token predictions are significantly more stable and do not require such adjustments.
As analyzed in Section~\ref{noprompt}, our $\mathcal{P}^\mathit{3}$ method allows for good classification scores without the need for a prompt.

\noindent
\textbf{Why are zero-shot classification capabilities not commonly used as an evaluation metric for large language models?}
Theoretically, zero-shot classification is an excellent, fair, and direct evaluation metric. However, the accuracy of next-token zero-shot classification is heavily influenced by the quality of the prompt. This reliance on prompt selection makes zero-shot classification a less reliable metric.
Conversely, using the accuracy distribution of subsequent tokens obtained by our proposed $\mathcal{P}^\mathit{3}$, which is much more stable and less affected by the prompt, provides a fairer and more reliable evaluation metric.

\section{Conclusion}

Prompt brittleness is a critical challenge in zero-shot classification, undermining the reliability and consistency of language model outputs. 
In this work, we proposed $ \mathcal{P}^\mathit{3} $, a novel method that introduces subsequent token predictions within a single model run to address this issue. 
Our extensive experiments across seven public benchmarks demonstrated that $ \mathcal{P}^\mathit{3} $  significantly mitigates prompt brittleness while achieving accuracy beyond the SoTA. 
Notably, $ \mathcal{P}^\mathit{3} $ performs equally well with or without prompts, greatly reducing the reliance on prompt engineering. 
These findings highlight $ \mathcal{P}^\mathit{3} $ ’s potential as an effective solution for robust zero-shot classification, making classification more efficient and reliable.



\section{Limitation}

In the paper, we discussed using existing tokens as placeholders <ph>. Training soft tokens as placeholders may be another solution. For auto-regressive language models, this approach involves training a soft token embedding of the ideal \texttt{<ph>} with a corpus. The formal expression of this token's properties is:
\begin{align}
\notag
\mathcal{LM}(\boldsymbol{x} \oplus \texttt{<ph>})_{y}
&= \sum_{v \in \mathcal{V}} \mathcal{LM}(\boldsymbol{x})_{v} \cdot \mathcal{LM}(\boldsymbol{x} \oplus v)_{y}
\end{align}
where \(\oplus\) denotes the concatenation of sequences.

Future work will focus on the behavior of tokens generated by our $ \mathcal{P}^\mathit{3} $ method across different types of models and languages.
Additionally, we plan to train a language model inherently equipped with \texttt{<ph>} tokens.

\bibliography{custom}

\appendix

\section{Appendix}
\label{sec:appendix}

\subsection{LM Output Examples}
\label{sec:OutputExamples}

Some examples of the language model's greedy output are in Table~\ref{tab:greedyOutputs}.
As shown, the generation of language models does not ensure that the next token will produce the discriminative words necessary for classification.

\begin{table}[h!]
\centering

\vspace{-0.3em}

\setlength{\abovecaptionskip}{0.15cm}  

\begin{tabular}{r|l}
\toprule
Input:&My ACL paper is accepted. It is \_ 
\\
Output:&\textcolor{blue}{a very} \textcolor{red}{good} paper. 
\\ \hline

Input:&My legs are broken. I am \_\\
Output: & \textcolor{blue}{in a lot of} \textcolor{red}{pain}. \\ \hline

Input:&I had my first job this year. It is \_ \\
Output:& \makecell[lt]{
\textcolor{blue}{a part-time job at a local restaurant.} \\
\textcolor{blue}{I am a waitress.} \\\textcolor{blue}{ I} \textcolor{red}{like}
 my job very much.
}
\\ \bottomrule
\end{tabular}

\caption{
Greedy output examples from LLaMa2-13B. \textcolor{blue}{Blue} text indicates non-discriminative expressions and \textcolor{red}{red} text indicates discriminative words for target categories \{positive, negative\}.
}

\label{tab:greedyOutputs}

  \vspace{-0.5em}

\end{table}

\subsection{Zero-Shot Classification: Worth Studying}
Zero-shot text classification remains a crucial research area, even as few-shot learning gains attention. Few-shot learning cannot fully replace zero-shot approaches, as it still relies on labeled data, which is not always feasible in many practical scenarios.

\textbf{Broader Applicability and Fewer Constraints}: Zero-shot learning is more aligned with real-world needs, especially when users cannot provide labeled examples, or when constructing even a single example for complex inputs is impractical. In contrast, few-shot learning assumes a shared distribution between provided and test examples, which is not always guaranteed.

\textbf{Core Foundation for Research}: Zero-shot learning serves as a more fundamental base for research, offering cleaner results without complexities like sample selection or ordering seen in few-shot studies, which can obscure core contributions.

\textbf{Greater challenge and a benchmark for intrinsic model capabilities}: Zero-shot classification directly tests the inherent capabilities of language models, free from the influence of examples. Our $ \mathcal{P}^\mathit{3} $ method further enhances the reliability of zero-shot evaluations by mitigating prompt instability, making it a more stable and valuable metric.

\subsection{Selected $\eta$}
The hyperparameter $\eta$ selections are in Table~\ref{tab:etatable}.

\begin{table}[!ht]
    \centering
    \begin{tabular}{l|cc}
    \toprule
        \textbf{Dataset} & \textbf{13B} & \textbf{70B}\\
        \midrule
        IMDb & 37 & 42 \\
        AGnews & 59 & 497\\
        Amazon & 4 & 17 \\
        DBpedia & 62 & 4 \\
        ISEAR & 8 & 5 \\
        SST-2 & 6 & 7 \\
        Yahoo & 65 & 4 \\

    \bottomrule
        
    \end{tabular}
    

\caption{
Hyperparameter settings for different datasets and models.
}
    
    \label{tab:etatable}
\end{table}
\label{sec:eta}

\subsection{Dataset Size}
The size of datasets we used in the paper is shown in Table~\ref{tab:datasize}.
\begin{table}[h!]
\centering
\begin{tabular}{ll}
\toprule
IMDb    & 25000  \\
AGnews  & 7600   \\
Amazon  & 400000 \\
DBpedia & 70000  \\
ISEAR   & 7666   \\
SST-2   & 1821   \\
Yahoo   & 60000 \\
\bottomrule
\end{tabular}
\caption{The size of datasets.}
\label{tab:datasize}
\end{table}

\subsection{Behavior Differences of Language Models}
\label{modeldiff}

\begin{figure*}[h!]
    \centering

    \includegraphics[width=1\textwidth]{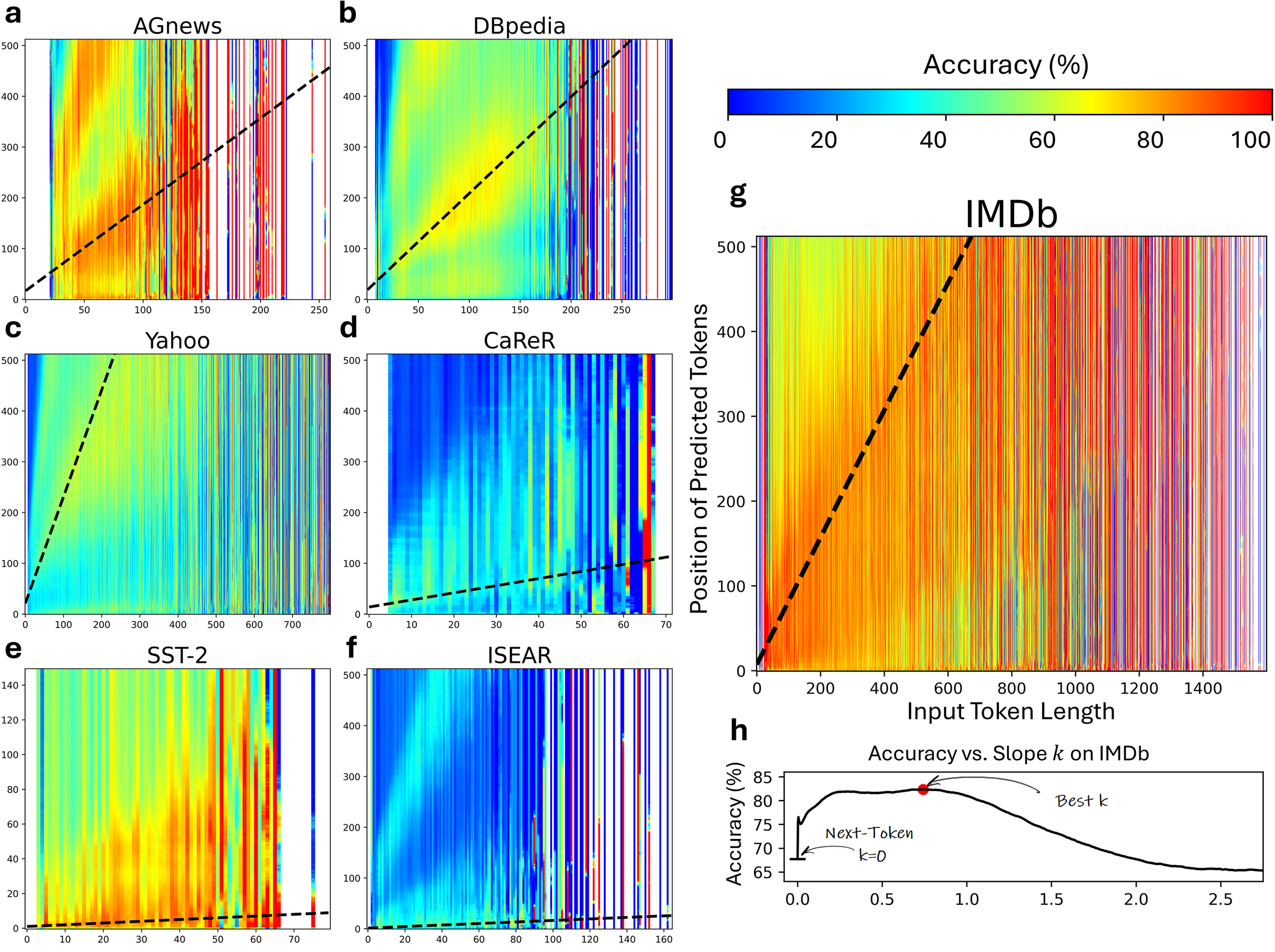}

\caption{
    \texttt{\textbf{abcdefg}}: Accuracy heatmaps for each dataset on 13b.
    In each subplot, the x-axis represents the text length $l_t$ (number of tokens in the text), and the y-axis represents the position of the token used for classification (0 represents the next-token).
    \texttt{\textbf{h}} shows the average accuracy of the IMDb dataset at different slopes \(k\) (with intercept \(b=-10\)). 
}

    \label{fig:avgheatmap}
\end{figure*}


Although, as discussed in Section 5.1, the use of the $ \mathcal{P}^\mathit{3} $ method consistently yields benefits, the classification performance of tokens at different positions varies.
In this regard, the 13B and 70B models exhibit different behaviors. Specifically, the performance of the 13B model is not only position-dependent but also influenced by the number of input tokens, whereas the 70B model does not exhibit this phenomenon.

As shown in Figure~\ref{fig:avgheatmap}, the distribution of well-performing (or poorly-performing) tokens in the 13B model follows a radial pattern, proportional to the number of input tokens. In contrast, the performance of tokens in the 70B model depends solely on their position and is independent of the input token length (exhibits vertical striping patterns).

In this paper, we treated it as an observed fact and leveraged it as a characteristic of language models. However, the exact underlying cause of this difference is challenging to determine due to the lack of transparency in LLaMA2’s training details.
We hypothesize that this distinction might arise from differences in the training process:

Stable positions in 70b: If the model underwent direct instruction-tuning or RLHF (Reinforcement Learning with Human Feedback) without distillation, the potential well-performing token positions would likely remain stable. This is because, for most classification tasks, answer lengths are consistent regardless of input variation.

Length-dependent positions in 13b: If the model underwent distillation from a larger model during instruction-tuning or RLHF, it might have inherited reasoning behaviors linked to input length, causing the observed correlation between input length and potential well-performing token positions.
We would like to interpret this result as a possible consequence of discrepancies in training processes and parameter scales. While we have not confirmed this hypothesis, we believe deeper exploration could lead to a more rigorous explanation. Since this behavior difference does not impact our primary claims regarding the mitigation of prompt brittleness, we treated it as an existing fact in our experiments.

The two models exhibit behavior differences, prompting us to adapt our method accordingly. For LLaMA2-13b, we observed a radial pattern in its plot, prompting us to set $\eta$ as the angle and use a line with slope $\tan(\eta)$ as our final result. For LLaMA2-70b, with a stable separability plot independent of input length, we define [0, $\eta$) as a fixed range for voting and applying calibration to enhance the results.

Furthermore, we found that using the votes of all 513 tokens obtained via $ \mathcal{P}^\mathit{3} $ method as pseudo-labels also produces patterns similar to those of the true labels. However, we did not pursue further investigation into this matter.

\begin{figure}[h!]
    \centering
    \includegraphics[width=1\linewidth]{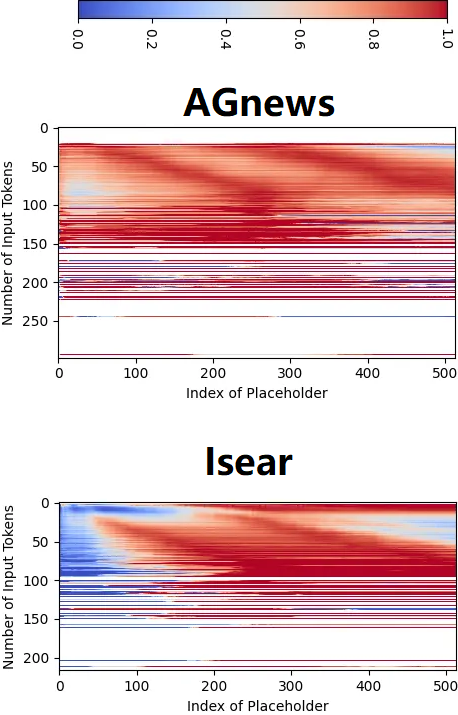}
    \caption{
Consistency plot with pseudo-labels, where the pseudo-labels are determined by the voting results of 513 tokens.
    }
\end{figure}

\subsection{FLOPs Estimation for Transformer Models}
\label{moreflops}

To estimate the FLOPs for the forward pass of a Transformer model, we consider the main computational components: attention and feed-forward networks (FFN).

For the attention mechanism, the input matrices \( Q \), \( K \), and \( V \) (with dimensions \( (B, s, h) \)) go through linear transformations to produce \( W_q \), \( W_k \), and \( W_v \):
\[
\text{QKV transformation FLOPs} = 6 B s h^2
\]
Next, the attention matrix is computed by multiplying the query matrix \( Q \) with the transposed key matrix \( K^T \), resulting in a matrix of size \( (s, s) \):
\[
\text{Attention matrix computation FLOPs} = 2 B s^2 h
\]
Then, the attention values are computed by multiplying the attention matrix with the value matrix \( V \):
\[
\text{Attention over values FLOPs} = 2 B s^2 h
\]
Finally, a linear projection is applied to the resulting matrix:
\[
\text{Post-attention linear projection FLOPs} = 2 B s h^2
\]
For the FFN, where the input dimension \( h \) is transformed to \( 4h \) and then back to \( h \), the FLOPs for this part are:
\[
\text{FFN FLOPs} = 16 B s h^2
\]
The language model head performs a linear transformation from the hidden size \( h \) to the vocabulary size \( V \), resulting in the following FLOPs:
\[
\text{LM head FLOPs} = 2 B s h V
\]
The total FLOPs per layer, considering attention and FFN, is the sum of the components above:
\[
\text{FLOPs per layer} = 24 B s h^2 + 4 B s^2 h
\]
Including the language model head, the total FLOPs per step for the forward pass in a model with \( l \) layers is:
\begin{align*}
\text{Total FLOPs per step} &= l \left( 24 B s h^2 + 4 B s^2 h \right) \\
&\quad + 2 B s h V
\end{align*}
Based on the formula, we also computed the case using the selected $\eta$s (Table~\ref{tab:datasetcomparison2}), as well as the case where all 512 placeholder tokens were used in our experiments (Table~\ref{tab:datasetcomparison3}).

\begin{table}[h!]
    \centering
    \resizebox{1\linewidth}{!}{
\setlength{\tabcolsep}{9pt} 
    \begin{tabular}{l|cc|cc}
    \toprule
        \multirow{2}{*}{\textbf{Dataset}} & \multicolumn{2}{c|}{\textbf{13B}} & \multicolumn{2}{c}{\textbf{70B}} \\
        \cmidrule{2-5}
        & \textbf{NTP} & $\boldsymbol{\mathcal{P}^\mathit{3}}$ & \textbf{NTP} & $\boldsymbol{\mathcal{P}^\mathit{3}}$  \\ \midrule
        Amazon & 2.82 & 3.00 & 14.26 & 16.47 \\ 
        IMDb & 8.31 & 14.73 & 41.91 & 47.41 \\ 
        AGnews & 1.62 & 4.31 & 8.20 & 73.32 \\
        DBpedia & 2.16 & 6.24 & 10.94 & 11.46 \\ 
        Yahoo & 3.67 & 11.78 & 18.56 & 19.08 \\ 
        SST-2 & 0.67 & 0.72 & 3.38 & 4.28 \\ 
        ISEAR & 0.70 & 0.79 & 3.57 & 4.22 \\ 
        \bottomrule
    \end{tabular}}
    \caption{
    Total Tflops calculation for the selected $\eta$.
    }
    \label{tab:datasetcomparison2}
\end{table}
\begin{table}[h]
    \centering
    \resizebox{1\linewidth}{!}{
\setlength{\tabcolsep}{9pt} 
    \begin{tabular}{l|cc|cc}
    \toprule
        \multirow{2}{*}{\textbf{Dataset}} & \multicolumn{2}{c|}{\textbf{13B}} & \multicolumn{2}{c}{\textbf{70B}} \\
        \cmidrule{2-5}
        & \textbf{NTP} & $\boldsymbol{\mathcal{P}^\mathit{3}}$ & \textbf{NTP} & $\boldsymbol{\mathcal{P}^\mathit{3}}$  \\ \midrule
        Amazon & 2.82 & 16.17 & 14.26 & 81.48 \\ 
        IMDb & 8.31 & 21.84 & 41.91 & 109.69 \\ 
        AGnews & 1.62 & 14.94 & 8.20 & 75.30 \\
        DBpedia & 2.16 & 15.50 & 10.94 & 78.09 \\ 
        Yahoo & 3.67 & 17.06 & 18.56 & 85.87 \\ 
        SST-2 & 0.67 & 13.96 & 3.38 & 70.37 \\ 
        ISEAR & 0.70 & 13.99 & 3.57 & 70.57 \\ 
        \bottomrule
    \end{tabular}}
    \caption{
    Total Tflops calculation for use of all 512 placeholder tokens in the experiment.
    }
    \label{tab:datasetcomparison3}
\end{table}

As can be seen, the overall computational cost is proportional to the number of tokens, indicating that our method does not impose a significant overhead.

\subsection{About Multi-Token Class Names}

A common challenge in zero-shot classification, particularly when dealing with class names that span multiple tokens, is how to calculate the class score from the prediction of a single token. This issue arises due to the way current language models generate and calibrate token probabilities.

In typical zero-shot approaches based on token generation probabilities, directly multiplying probabilities for individual tokens within a sequence often fails to yield reliable classification scores. This problem is particularly pronounced when class descriptions consist of multiple tokens. Some studies address this challenge by integrating external resources, such as knowledge bases or validation datasets, which can help identify relevant tokens or synonyms associated with each class. For example, one common method is to sum the probabilities of all relevant tokens (e.g., "good" and "positive") to compute a cumulative score. However, such methods often require extensive preprocessing, which can limit their applicability in more dynamic or real-time scenarios.

In this work, we focus on exploring the robustness of subsequent token predictions. To ensure consistency with previous research and simplify the setup, we adopted the widely used practice of mapping each class to a single token (e.g., "good" for the positive class and "bad" for the negative class), as seen in standard datasets.

An interesting aspect of our $ \mathcal{P}^\mathit{3} $ method, however, is its ability to naturally handle multi-token class names. By aggregating probabilities across subsequent tokens, $ \mathcal{P}^\mathit{3} $ can provide a more coherent approach to multi-token class labels. Rather than relying on the prediction of a single token, $ \mathcal{P}^\mathit{3} $ aggregates predictions across tokens, enabling it to better align with natural language. For instance, in cases where tokens like "ppi" or "ness" are unlikely to appear in the immediate context, $ \mathcal{P}^\mathit{3} $ can still effectively aggregate across related tokens, such as "good" and "positive," to compute a more accurate class score.

This approach presents an important advantage over traditional methods, offering a more scalable and flexible framework for zero-shot classification tasks, particularly in scenarios involving complex or multi-token class names.

\subsection{Position Plot}
\begin{figure}[h!]
    \centering

    \includegraphics[width=1\linewidth]{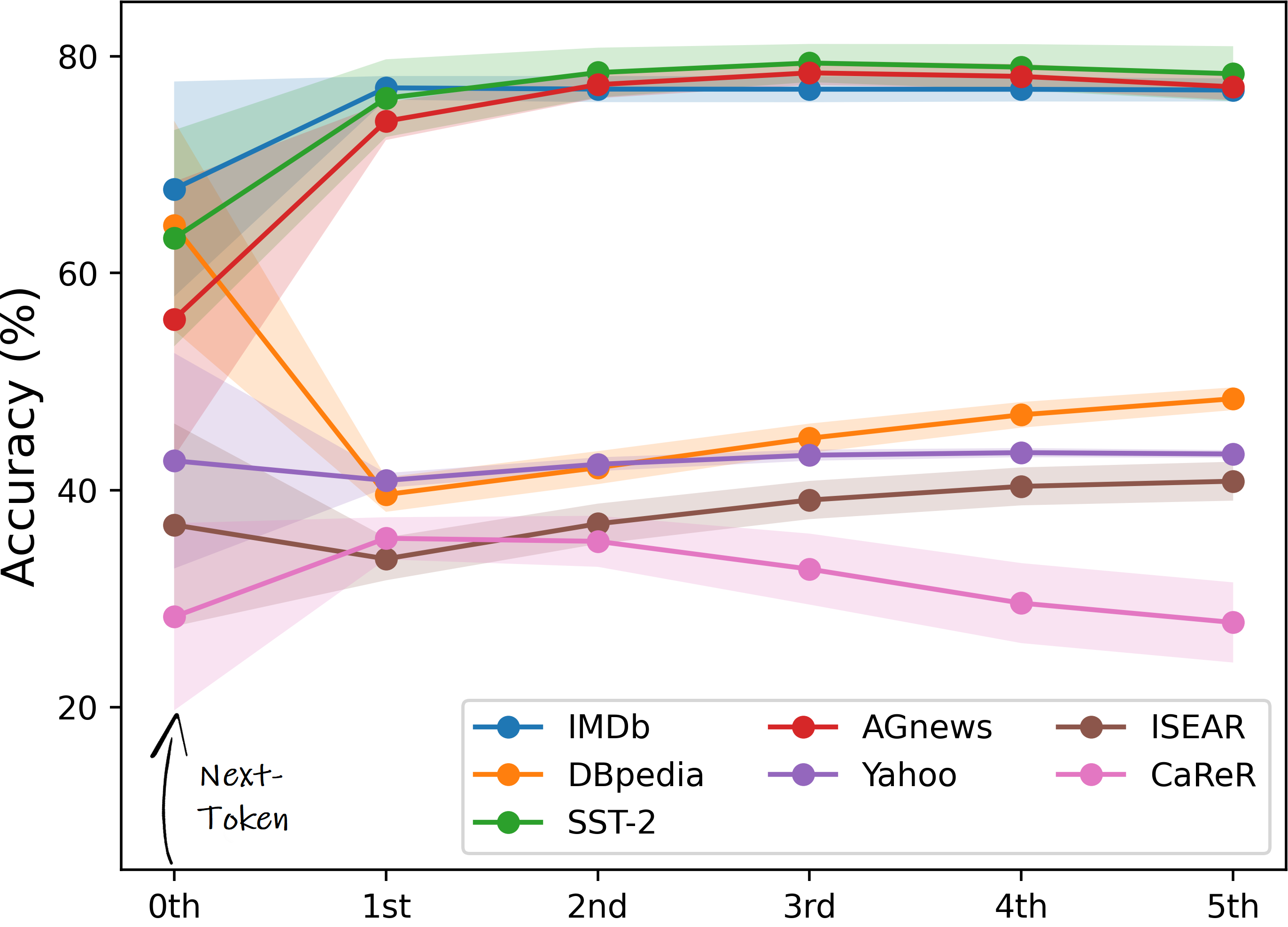}
    \caption{
    Accuracy and cross-prompt standard deviation for each dataset using predicted tokens at fixed positions on 13b. The x-axis shows token position. Dots indicate average accuracy and shaded areas represent standard deviation.
    }
    \label{fig:fix-pos}
\end{figure}

\subsection{Prompts used in the experiment.}
\label{promptlist}
To evaluate the effectiveness and robustness of our Placeholding Parallel Prediction method, we conduct extensive experiments on seven datasets with various kinds of prompts, including empty prompts. The specific prompt settings for different datasets are shown in Table 10-16.

\begin{table*}[]
\centering
\begin{tabular}{r|p{12.5cm}}
\toprule
ID  & Template \\
\toprule
1   & {title}\textbackslash n{text}\textbackslash nMy feedback to it is \\
2   & {title}\textbackslash n{text}\textbackslash nOverall, my feedback to it is \\
3   & {title}\textbackslash n{text}\textbackslash nAfter considering all aspects, my feedback to it is \\
4   & {title}\textbackslash n{text}\textbackslash nAfter considering all aspects, my viewpoint is \\
5   & {title}\textbackslash n{text}\textbackslash nReflecting on the above, my viewpoint is \\
6   & {title}\textbackslash n{text}\textbackslash nOverall, my perspective on it is \\
7   & {title}\textbackslash n{text}\textbackslash nOverall, my takeaway is \\
8   & {title}\textbackslash n{text}\textbackslash nIn summary, I would say \\
9   & {title}\textbackslash n{text}\textbackslash nConsidering the details provided, my emotional reaction is \\
10  & {title}\textbackslash n{text}\textbackslash nTaking into account the experience shared, my viewpoint is \\
11  & {title}\textbackslash n{text}\textbackslash nReflecting on the content, my emotional stance is \\
12  & {title}\textbackslash n{text}\textbackslash nGiven the information above, my perspective on it is \\
13  & {title}\textbackslash n{text}\textbackslash nAnalyzing the feedback, my emotional assessment is \\
14  & {title}\textbackslash n{text}\textbackslash nBased on the review, my overall sentiment impression is \\
15  & {title}\textbackslash n{text}\textbackslash nWeighing up the insights, my sentiment conclusion is \\
16  & {title}\textbackslash n{text}\textbackslash nAfter thoroughly considering the review, my sentiment perspective is \\
17  & Text: {title} {text}\textbackslash nSentiment: \\
18  & Text: {title} {text}\textbackslash nSentiment Analysis: The overall sentiment is \\
19  & {title}\textbackslash n{text}\textbackslash nAll in all, it was \\
20  & {title}\textbackslash n{text}\textbackslash nIn summary, it was \\
21  & {title}\textbackslash n{text}\textbackslash nIn essence, it was \\
22  & {title}\textbackslash n{text}\textbackslash nIn conclusion, it was \\
23  & {title}\textbackslash n{text}\textbackslash nTo sum up, it's \\
24  & {title}\textbackslash n{text} All in all \\
25  & {title}\textbackslash n{text} Just \\
26  & {title}\textbackslash n{text} It was \\
27  & {title}\textbackslash n{text} It is \\
28  & {title}\textbackslash n{text} That is \\
29  & {title}\textbackslash n{text} That's \\
30  & {title}\textbackslash n{text} But it is \\
31  & {title}\textbackslash n{text} \\
\bottomrule
\end{tabular}
\caption{Prompt list for Amazon Review Polarity dataset.}
\end{table*}

\begin{table*}[]
\centering
\begin{tabular}{r|p{12.5cm}}
\toprule
ID  & Template \\
\toprule
1   & [text] My feedback to the film is \\
2   & [text] Overall, my feedback to the film is \\
3   & [text] After considering all aspects, my feedback to the film is \\
4   & [text] After considering all aspects, my viewpoint is \\
5   & [text] Reflecting on the above, my viewpoint is \\
6   & [text] Overall, my perspective on the film is \\
7   & [text] Overall, my takeaway is \\
8   & [text] In summary, I would say \\
9   & [text] I think it is \\
10  & [text] Overall, I think it is \\
11  & [text] Considering everything, my feedback is \\
12  & [text] Considering everything, I think it is \\
13  & [text] After thinking about it, my feedback is \\
14  & [text] Overall, I see it as \\
15  & [text] Taking all factors into account, my assessment of it is \\
16  & [text] Considering the details provided, my emotional reaction is \\
17  & [text] Taking into account the experience shared, my sentiment is \\
18  & [text] Reflecting on the content, my emotional stance is \\
19  & [text] Given the information above, my sentiment evaluation is \\
20  & [text] Analyzing the feedback, my emotional assessment is \\
21  & [text] Based on the review, my overall sentiment impression is \\
22  & [text] Weighing up the insights, my sentiment conclusion is \\
23  & [text] After thoroughly considering the review, my sentiment perspective is \\
24  & Text: [text] Sentiment: \\
25  & Text: [text] Sentiment Analysis: The overall sentiment is \\
26  & [text] All in all, the film was \\
27  & [text] In summary, the film was \\
28  & [text] In essence, the film was \\
29  & [text] In conclusion, the film was \\
30  & [text] To sum up, the film was \\
31  & [text] All in all \\
32  & [text] Just \\
33  & [text] It was \\
34  & [text] It is \\
35  & [text] That is \\
36  & [text] That's \\
37  & [text] But it is \\
38  & [text] \\
\bottomrule
\end{tabular}
\caption{Prompt list for IMDb dataset.}
\end{table*}

\begin{table*}[]
\centering
\begin{tabular}{r|p{12.5cm}}
\toprule
ID  & Template \\
\toprule
1   & [text] My feedback to the film is \\
2   & [text] Overall, my feedback to the film is \\
3   & [text] After considering all aspects, my feedback to the film is \\
4   & [text] After considering all aspects, my viewpoint is \\
5   & [text] Reflecting on the above, my viewpoint is \\
6   & [text] Overall, my perspective on the film is \\
7   & [text] Overall, my takeaway is \\
8   & [text] In summary, I would say \\
9   & [text] I think it is \\
10  & [text] Overall, I think it is \\
11  & [text] Considering everything, my feedback is \\
12  & [text] Considering everything, I think it is \\
13  & [text] After thinking about it, my feedback is \\
14  & [text] Overall, I see it as \\
15  & [text] Taking all factors into account, my assessment of it is \\
16  & [text] Considering the details provided, my emotional reaction is \\
17  & [text] Taking into account the experience shared, my sentiment is \\
18  & [text] Reflecting on the content, my emotional stance is \\
19  & [text] Given the information above, my sentiment evaluation is \\
20  & [text] Analyzing the feedback, my emotional assessment is \\
21  & [text] Based on the review, my overall sentiment impression is \\
22  & [text] Weighing up the insights, my sentiment conclusion is \\
23  & [text] After thoroughly considering the review, my sentiment perspective is \\
24  & Text: [text] Sentiment: \\
25  & Text: [text] Sentiment Analysis: The overall sentiment is \\
26  & [text] All in all, the film was \\
27  & [text] In summary, the film was \\
28  & [text] In essence, the film was \\
29  & [text] In conclusion, the film was \\
30  & [text] To sum up, the film was \\
31  & [text] All in all \\
32  & [text] Just \\
33  & [text] It was \\
34  & [text] It is \\
35  & [text] That is \\
36  & [text] That's \\
37  & [text] But it is \\
38  & [text] \\
\bottomrule
\end{tabular}
\caption{Prompt list for SST-2 dataset.}
\end{table*}

\begin{table*}[]
\centering
\begin{tabular}{r|p{12.5cm}}
\toprule
ID  & Template \\
\toprule
1   & [title] [text] This topic is about \\
2   & [title] [text] The label that best describes this news article is \\
3   & [title] [text] This piece of news is regarding \\
4   & [title] [text] The news article is about \\
5   & [title] [text] Central themes of this news piece encompass \\
6   & [title] [text] The central theme of this article revolves around \\
7   & [title] [text] It can be labeled as \\
8   & [title] [text] Its category is \\
9   & [title] [text] In this article, it talks about \\
10  & [title] [text] The content is a kind of \\
11  & [title] [text] I think the news can be classified as \\
12  & [title] [text] I would classify it as \\
13  & [title] [text] Based on the description, its category is \\
14  & [title] [text] In this context, the content falls into the category of \\
15  & Text: [title] [text] Category: \\
16  & Text: [title] [text] Topic Classification: The overall topic is \\
17  & [title] [text] All in all, it was \\
18  & [title] [text] In summary, it was \\
19  & [title] [text] In essence, it was \\
20  & [title] [text] In conclusion, it was \\
21  & [title] [text] To sum up, it's \\
22  & [title] [text] All in all \\
23  & [title] [text] Just \\
24  & [title] [text] It was \\
25  & [title] [text] It is \\
26  & [title] [text] That is \\
27  & [title] [text] That's \\
28  & [title] [text] But it is \\
29  & [title] [text] \\
\bottomrule
\end{tabular}
\caption{Prompt list for AGnews dataset.}
\end{table*}

\begin{table*}[]
\centering
\begin{tabular}{r|p{12.5cm}}
\toprule
ID  & Template \\
\toprule
1   & [title] [text] The category of [title] is \\
2   & [title] [text] The label that best describes [title] is \\
3   & [title] [text] So, [title] is \\
4   & [title] [text] In this sentence, [title] is \\
5   & [title] [text] [title] is a kind of \\
6   & [title] [text] [title] can be classified as \\
7   & [title] [text] [title] is an example of \\
8   & [title] [text] [title] belongs to \\
9   & [title] [text] I think [title] is \\
10  & [title] [text] I would classify [title] as \\
11  & [title] [text] Based on the description, its category is \\
12  & [title] [text] In this context, [title] falls into the category of \\
13  & Text: [title] [text] Category: \\
14  & Text: [title] [text] Topic Classification: The overall topic is \\
15  & [title] [text] All in all, it is \\
16  & [title] [text] In summary, it is \\
17  & [title] [text] In essence, it is \\
18  & [title] [text] In conclusion, it is \\
19  & [title] [text] To sum up, it's \\
20  & [title] [text] All in all \\
21  & [title] [text] Just \\
22  & [title] [text] It was \\
23  & [title] [text] It is \\
24  & [title] [text] That is \\
25  & [title] [text] That's \\
26  & [title] [text] But it is \\
27  & [title] [text] \\
\bottomrule
\end{tabular}
\caption{Prompt list for DBpedia dataset.}
\end{table*}

\begin{table*}[]

\begin{tabular}{r|l}
\toprule
ID.  & Template \\
\midrule
1    & [title] [text] This topic is about \\
2         & [title] [text] The label that best describes this question is \\
3        & [title] [text] This issue is regarding \\
4         & [title] [text] This discussion is about \\
5        & [title] [text] This discussion is regarding \\
6         & [title] [text] This issue is about \\
7          & [title] [text] The label that best describes this issue is \\
8          & [title] [text] I would classify this question as \\
9          & [title] [text] It can be labeled as \\
10        & [title] [text] Overall, The most fitting category for this issue is \\
11         & [title] [text] The content is associated with \\
12        & [title] [text] I think it belongs to \\
13         & [title] [text] I would classify it as \\
14         & [title] [text] This issue falls into the category of \\

15  & Text: [title] [text] Category: \\
16        & Text: [title] [text] Topic Classification: The overall topic is \\

17    & [title] [text] All in all, it was \\
18        & [title] [text] In summary, it was \\
19        & [title] [text] In essence, it was \\
20        & [title] [text] In conclusion, it was \\
21        & [title] [text] To sum up, it's \\

22   & [title] [text] All in all \\
23        & [title] [text] Just \\
24        & [title] [text] It was \\
25        & [title] [text] It is \\
26        & [title] [text] That is \\
27       & [title] [text] That's \\
28         & [title] [text] But it is \\

29  & [title] [text] \\
\bottomrule
\end{tabular}
\caption{Prompt list for Yahoo dataset.}
\end{table*}

\begin{table*}[]
\centering
\begin{tabular}{r|p{12.5cm}}
\toprule
ID  & Template \\
\toprule
1   & [text] In summary, I would say \\
2   & [text] I think it is \\
3   & [text] Overall, I think it is \\
4   & [text] Considering everything, I think it is \\
5   & [text] Overall, I see it as \\
6   & [text] In summary, I would say \\
7   & [text] I feel \\
8   & [text] Overall, I feel \\
9   & [text] Overall, my feeling towards it is \\
10  & [text] This text expresses \\
11  & [text] It is a feeling of \\
12  & [text] The sentiment is \\
13  & [text] It is \\
14  & [text] This conveys a sense of \\
15  & [text] I am \\
16  & [text] The overall impression is \\
17  & [text] From my perspective, it is \\
18  & [text] In my view, the feeling is \\
19  & [text] This passage makes me feel \\
20  & [text] It seems to evoke a feeling of \\
21  & [text] This text primarily conveys \\
22  & [text] From this, I sense an emotion of \\
23  & [text] It can be interpreted as expressing \\
24  & [text] The underlying emotion seems to be \\
25  & [text] This narrative elicits \\
26  & [text] Feeling-wise, this comes across as \\
27  & [text] This evokes \\
28  & [text] The emotional tone here is \\
29  & [text] This story is imbued with \\
30  & [text] One could interpret this as \\
31  & [text] This text leaves the impression of \\
32  & Text: [text] Emotion: \\
33  & Text: [text] Emotion Recognition: The overall emotion is \\
34  & [text] All in all, it was \\
35  & [text] In summary, it was \\
36  & [text] In essence, it was \\
37  & [text] In conclusion, it was \\
38  & [text] To sum up, it was \\
39  & [text] All in all \\
40  & [text] Just \\
41  & [text] It was \\
42  & [text] It is \\
43  & [text] That is \\
44  & [text] That's \\
45  & [text] But it is \\
46  & [text] \\
\bottomrule
\end{tabular}
\caption{Prompt list for ISEAR dataset.}
\end{table*}

\end{document}